\documentclass[twocolumn]{wlscirep}
\usepackage[utf8]{inputenc}
\usepackage[T1]{fontenc}
\usepackage{subfig}
\usepackage{url}

\title{High carbon stock mapping at large scale with optical satellite imagery and spaceborne LIDAR}

\author[1,*]{Nico Lang}
\author[1]{Konrad Schindler}
\author[1,2]{Jan Dirk Wegner}
\affil[1]{EcoVision Lab, Photogrammetry and Remote Sensing, ETH Zurich, Switzerland}
\affil[2]{Institute for Computational Science, University of Zurich, Switzerland}

\affil[*]{nico.lang@geod.baug.ethz.ch}


\begin{abstract}
The increasing demand for commodities is leading to changes in land use worldwide. In the tropics, deforestation, which causes high carbon emissions and threatens biodiversity, is often linked to agricultural expansion. While the need for deforestation-free global supply chains is widely recognized, making progress in practice remains a challenge. Here, we propose an automated approach that aims to support conservation and sustainable land use planning decisions by mapping tropical landscapes at large scale and high spatial resolution following the High Carbon Stock (HCS) approach. A deep learning approach is developed that estimates canopy height for each 10~m Sentinel-2 pixel by learning from sparse GEDI LIDAR reference data, achieving an overall RMSE of 6.3~m. We show that these wall-to-wall maps of canopy top height are predictive for classifying HCS forests and degraded areas with an overall accuracy of 86~\% and produce a first high carbon stock map for Indonesia, Malaysia, and the Philippines.
\end{abstract}

\begin{document}

\flushbottom
\maketitle

\thispagestyle{empty}

As the world's population grows, the demand for food, timber, and other commodities continues to rise, and with it the demand for agricultural land\cite{united2019world,valin2014future}. It is estimated that almost a third of the global land has been transformed within the last six decades\cite{winkler2021global}.
Although land use changes are taking place worldwide, tropical forests receive particular attention because their transformation causes high ecological damage. Global gross forest fluxes between 2001 and 2019 are dominated by deforestation in Tropical and subtropical forests, accounting for 78~\% of gross emissions ($6.3\pm2.4$~GtCO2e~yr$^{-1}$)\cite{harris2021global}. 
While being one of the most important terrestrial carbon stocks and among the richest biodiversity ecosystems on land\cite{scheffers2012we}, the pressure on tropical forests keeps growing\cite{karger2021limited,hoang2021mapping}.
Assessing the drivers of deforestation in the tropics is challenging, but there is evidence that industrial agriculture is the main driver of deforestation in the tropics\cite{jayathilake2021drivers,austin2019causes,wicke2011exploring,gaveau2016rapid}. Through the export of these tropical commodities tropical deforestation is linked to the global supply chain\cite{action2019stepping,hoang2021mapping}, making this a pressing global sustainability challenge.

The high carbon stock approach (HCSA)\footnote{\href{http://highcarbonstock.org/}{highcarbonstock.org} (2021-06-10)} is a methodology accepted by NGOs and supported by companies that produce, trade, and process commodities and that are committed to reduce deforestation associated with their supply chain. The HCSA aims at conserving high carbon stock forests by guiding the development in tropical countries in a more sustainable way. Its initial focus has been the palm oil and pulpwood sectors in tropical Asia and the HCSA is integrated in the Roundtable on Sustainable Palm Oil (RSPO) certification since November 2018\cite{hcsa2020rspo}. 
The toolkit elaborated by the HCSA serves as a landuse planning tool and is designed to account for all important factors to protect primary rain forest, while ensuring land use rights of local communities\cite{rosoman2017hcs}.
A landscape stratification has been defined that is mainly based on the carbon stock or in other words the aboveground biomass (hereafter "biomass") stored in forests\footnote{Aboveground biomass consists of $\approx$47\% carbon\cite{mcgroddy2004scaling}, a value also adopted for the Intergovernmental Panel on Climate Change (IPCC) guidelines\cite{eggleston20062006}.}. 
While a full assessment includes consideration of species composition, canopy closure, and successional state, indicative HCS mapping is based on a set of carbon density thresholds that categorize the landscape into high carbon stock (HCS) forests and degraded lands such as open land and scrub (OLS). The former must be protected, while the latter may be considered for further economical development, for instance to extend existing plantations. This strategy shall reduce the carbon emissions and ecological damage caused by land use changes.

Up until now HCS forest mapping is costly, because targeted airborne laser scanning (ALS) missions must be carried out to measure forest structure parameters like canopy height and cover, from which the carbon density is then derived either via field plot calibration or with pre-calibrated allometric equations. 
At an estimated\cite{rejou2019upscaling} 200--500 Euro per km$^{2}$, mapping the entire land surface of Indonesia, Malaysia, and Philippines ($\approx$2.5 million km$^{2}$)\cite{unstats} with ALS would amount to 0.5--1.2 billion Euro.
In practice, scanning campaigns are thus restricted to a regional scale and are updated infrequently or not at all, to reduce costs. Consequently, HCS mapping is typically based on semi-automatic analyses of satellite images, which affects the quality of the resulting maps because (i) simple regression models do not adequately capture the relationship between satellite image pixels and biomass, and (ii) the visual interpretation of satellite images in terms of carbon density is difficult for human operators and prone to subjective biases. Moreover, field plots on the ground remain an important component of biomass mapping, providing calibration and validation data for any of the mentioned sensing technologies. The associated field work is, however, time-consuming and does not scale to large regions. In summary, there is a need for a highly automated, objective approach to map carbon density and indicative\footnote{"Indicative" refers to maps based only on measurable information, before accounting for political factors such as traditional land rights.} HCS with high spatial resolution and at large scale. 

Measuring biomass and monitoring carbon stock is key for a host of Earth science questions, and also to assess actions and commitments to reduce CO$_2$ emissions. But large scale biomass estimation (e.g., from remote sensing data) remains challenging, especially in the tropics, and is an active scope of research\cite{rodriguez2017quantifying,mitchard2013uncertainty,mitchard2014markedly}. Important limitations are the still imperfect physical understanding of how data from several sensors correlate with biomass, but also the scarcity of calibration data, since collecting biomass data on the ground is itself a formidable task. To date, a systematically collected global database of aboveground biomass reference data is lacking\cite{duncanson2021aboveground}.
The lack of data to train advanced, data-hungry statistical models has complicated biomass prediction from satellite data. Recently, the data situation, and with it the potential of powerful computational tools like deep learning, has improved through NASA's GEDI mission, a space based laser scanner on board the International Space Station that has been collecting sparse, globally well-distributed measurements of vertical vegetation structure since March 2019\cite{dubayah2020global}. The GEDI lidar sensor has been designed for the retrieval of biomass, and among the large-scale forest structure variables that can be derived from lidar data the single most important predictor of biomass is canopy height\cite{jucker2017allometric,dubayah2010estimation,saatchi2011benchmark,baccini2012estimated,asner2014mapping,silva2018comparison,qi2019forest,drake2002estimation}.

We propose a deep learning approach that utilises publicly available Sentinel-2 satellite imagery from the European Space Agency (ESA) in combination with data from NASA's GEDI lidar mission to estimate canopy top height at a ground sampling distance (GSD) of 10~m. We then demonstrate that this canopy top height map is predictive for classifying HCS forests in tropical Asia and produce an \textit{indicative} high carbon stock map for the three countries: Indonesia, Malaysia, and Philippines.
In a first step sparse canopy top height estimates from GEDI full waveform data\cite{lang2021global} are fused with ESA's Sentinel-2 optical images to create a wall-to-wall map of canopy top height. A deep convolutional neural network (CNN) is trained to estimate a canopy top height for every 10~m$\times$10~m pixel of Sentinel-2\cite{lang2019country}. 
In a second step the carbon density map is regressed from the canopy top height map with an ensemble of small CNNs that is trained on carbon density data from an airborne lidar campaign in Borneo Sabah\cite{asner2018mapped}. The last step then applies the carbon thresholds defined by the HCSA, and overlays the HCS classification with additional map layers to explicitly identify tall crops\cite{rodriguez2021mapping,rodriguez2021coconut} (i.e., oil palm and coconut plantations) and urban regions\cite{marcel_buchhorn_2020_3939050}.

The canopy top height and indicative high carbon stock maps are available at: \href{http://doi.org/10.5281/zenodo.5012448}{doi.org/10.5281/zenodo.5012448}\cite{lang_2021_5012448}, and can be interactively explored in the Google Earth Engine App: \href{https://nlang.users.earthengine.app/view/canopy-height-and-carbon-stock-southeast-asia-2020}{nlang.users.earthengine.app/view/canopy-height-and-carbon-stock-southeast-asia-2020}.

\section*{Results}
\subsection*{Canopy top height mapping}
\begin{figure*}
    \centering
    \subfloat[]{{ \includegraphics[width=0.4\textwidth]{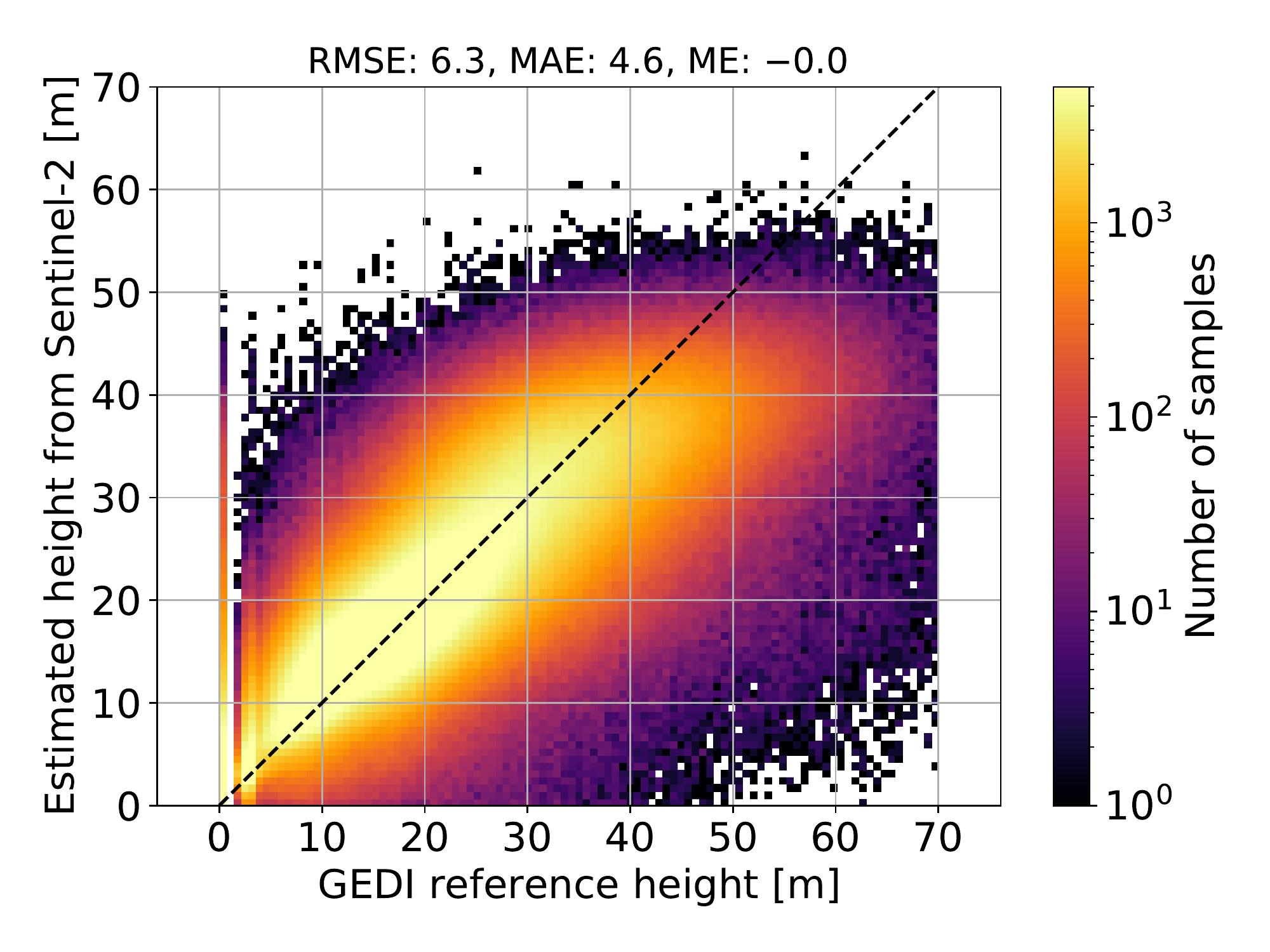} }}%
    \subfloat[]{{ \includegraphics[width=0.4\textwidth]{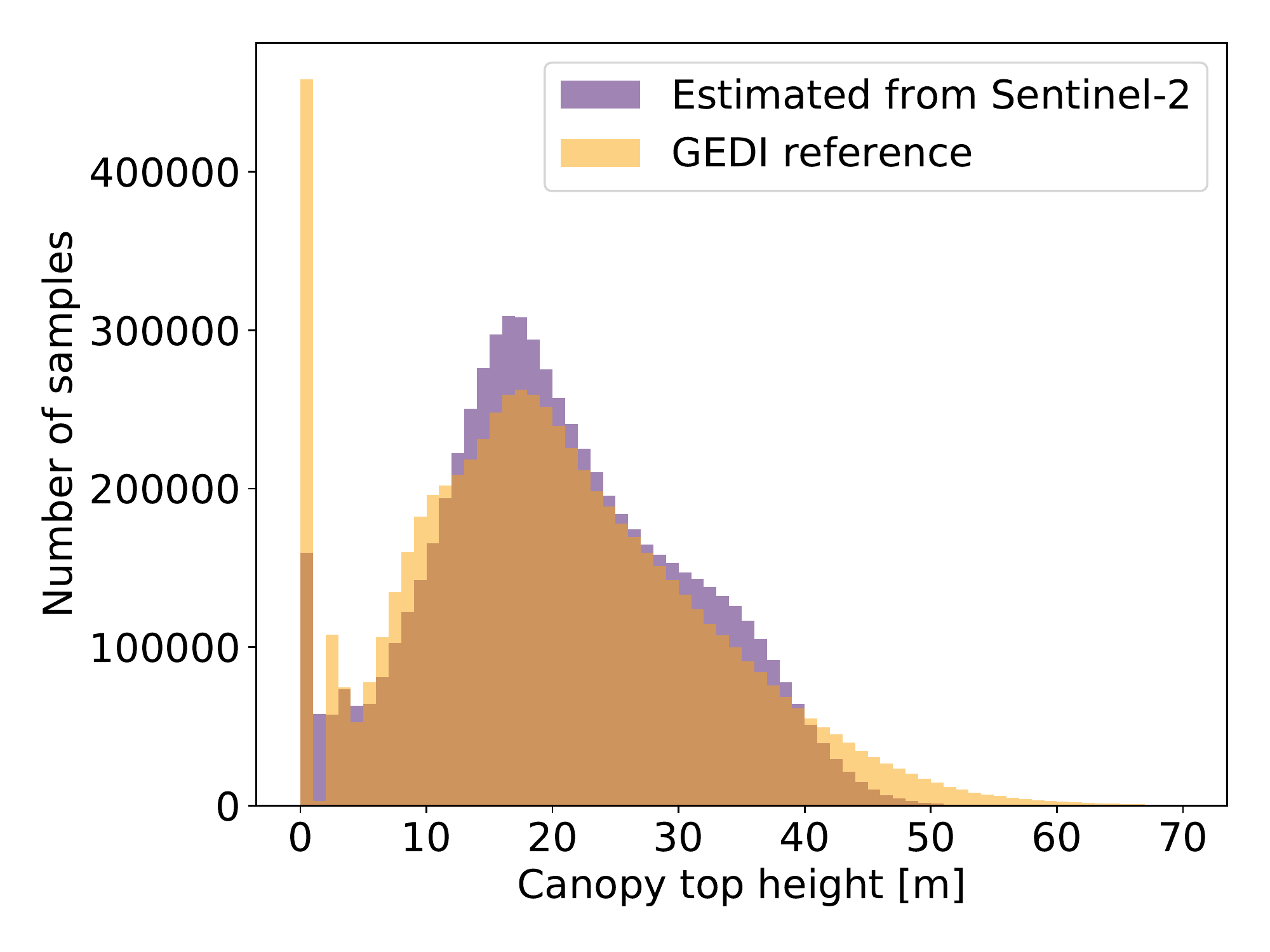} }}
    
    \caption{Canopy top height estimation from Sentinel-2 images. a) Confusion plot with GEDI reference data versus prediction from Sentinel-2. b) Marginal distribution of predicted and reference canopy top heights.} 
    \label{fig:canopy_results}
\end{figure*}

To evaluate the canopy top height regression in Southeast Asia, we hold-out a random subset of Sentinel-2 tiles and its corresponding GEDI canopy top height data. The total of 92 validation regions, each with an approximate area of 100~km$\times$100~km, are distributed uniformly over Southeast Asia. 
The performance of the CNN estimates against the unseen GEDI data yields an RMSE of 6.3~m and a MAE of 4.6~m with no overall bias. Low vegetation is slightly overestimated and tall canopies are underestimated (Fig.~\ref{fig:canopy_results}). Overall, the GEDI canopy heights and the estimated heights from Sentinel-2 follow a similar marginal distribution (Fig.~\ref{fig:canopy_results}b). 
The canopy height estimation saturates around 50~m, which is comparable to what has been observed when training on regional airborne lidar data\cite{lang2019country}. 
The zero height class (corresponding to non-vegetated land, which was set using the Sentinel-2 L2A scene classification) is frequently overestimated. These errors may occur at class boundaries between vegetation and non-vegetated areas, where both errors in the L2A scene classification, errors in the geolocation of GEDI validation footprints, as well as errors in the estimated canopy height aggregate in the evaluation.

\subsection*{Estimating carbon stock from canopy top height}
\begin{figure*}
    \centering
    \subfloat[]{{ \includegraphics[width=0.4\textwidth]{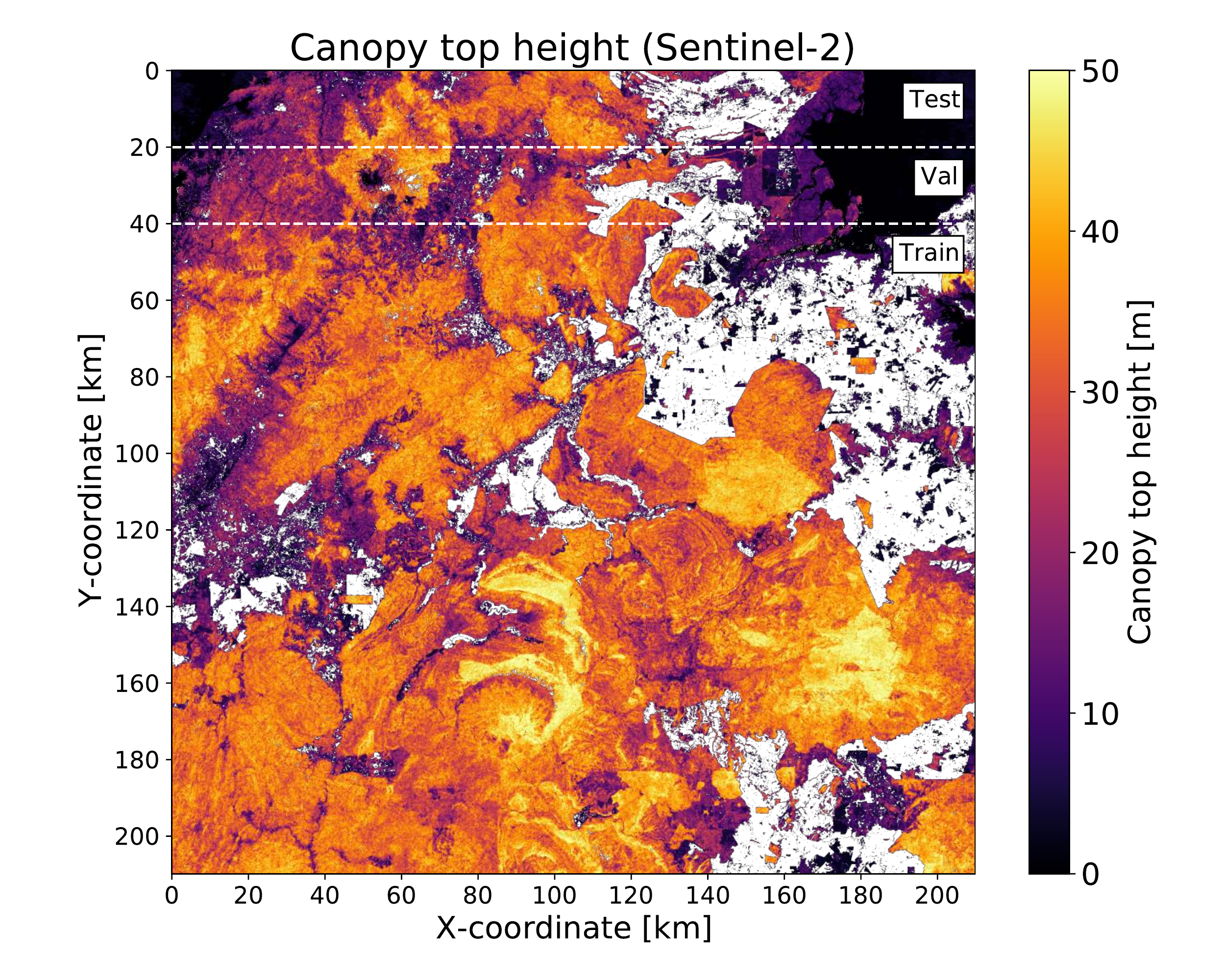} }}%
    \subfloat[]{{ \includegraphics[width=0.4\textwidth]{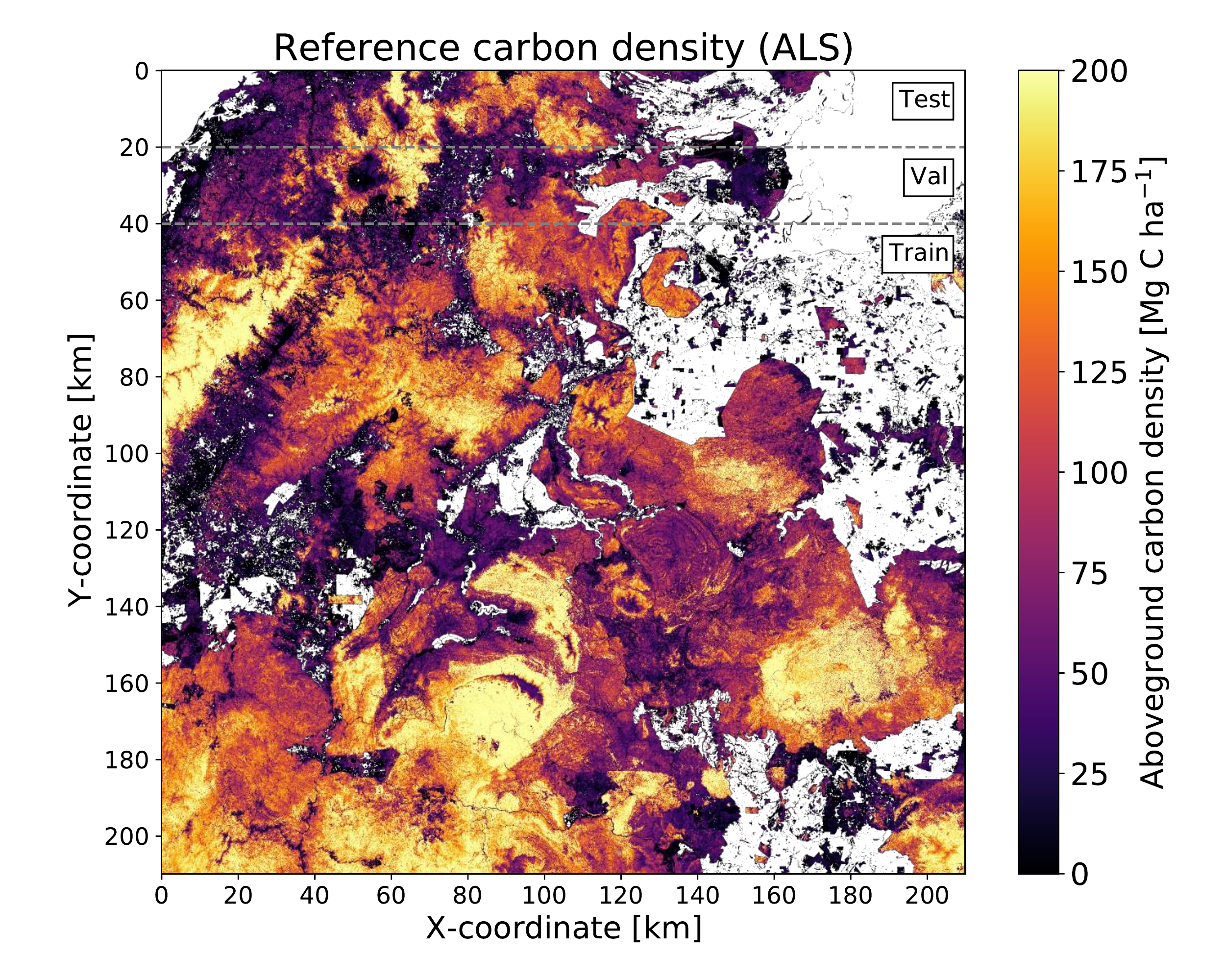} }}
    
    \subfloat[]{{ \includegraphics[width=0.4\textwidth]{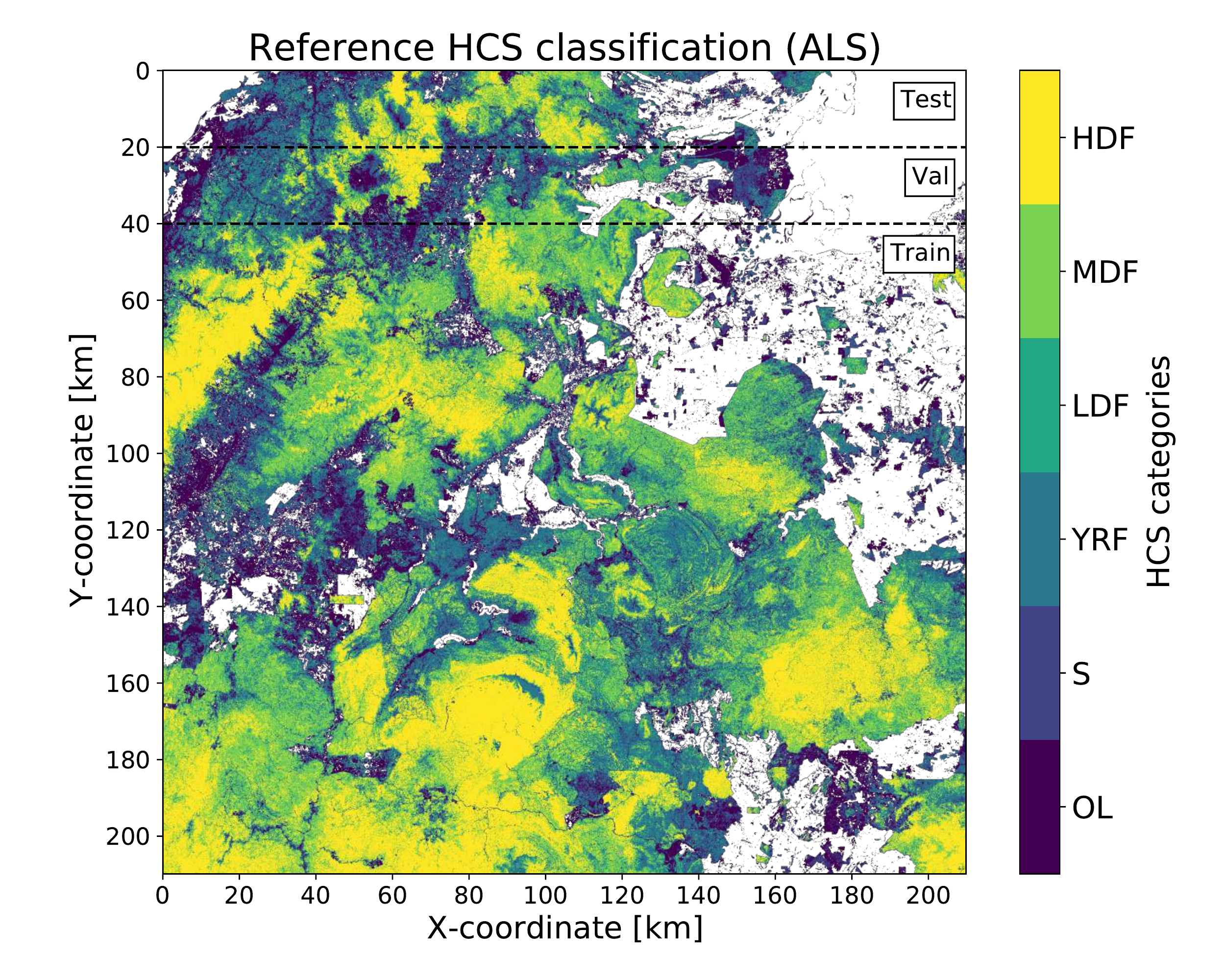} }}%
    \subfloat[]{{ \includegraphics[width=0.4\textwidth]{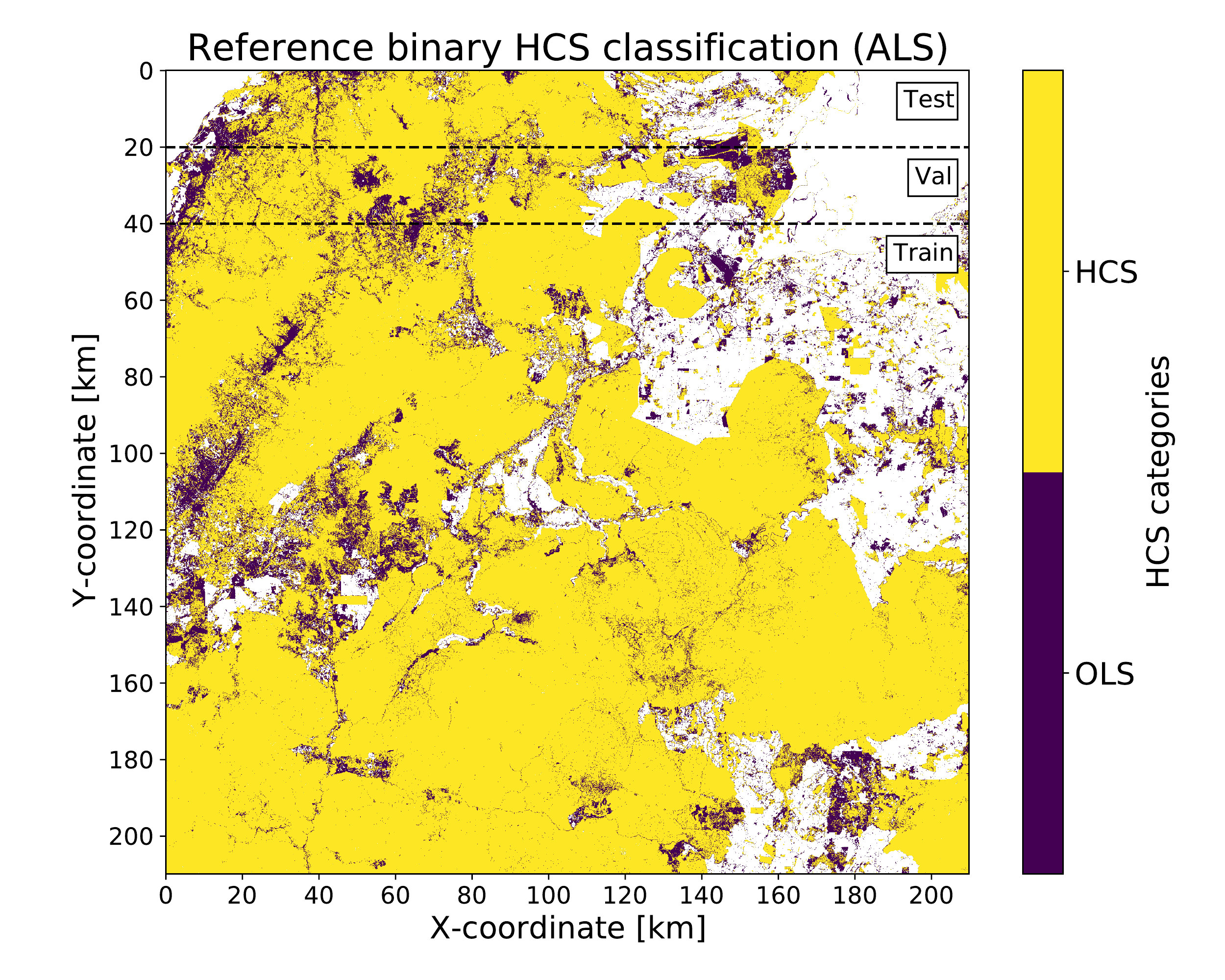} }}
    
    \caption{Calibration site with carbon density from an airborne lidar campaign (ALS) in Sabah, Borneo Malaysia\cite{asner2018mapped, asner_gregory_p_2021_4549461}. a) Canopy top height estimates from Sentinel-2. b) Aboveground carbon stock reference data from ALS\cite{asner2018mapped,asner_gregory_p_2021_4549461} c,d) HCS classification derived from ALS carbon density. Plantations have been masked out by thresholding existing palm tree density maps\cite{rodriguez2021mapping}.} 
    \label{fig:als_region}
\end{figure*}

\begin{figure*}
    \centering
    \subfloat[]{{ \includegraphics[width=0.6\textwidth]{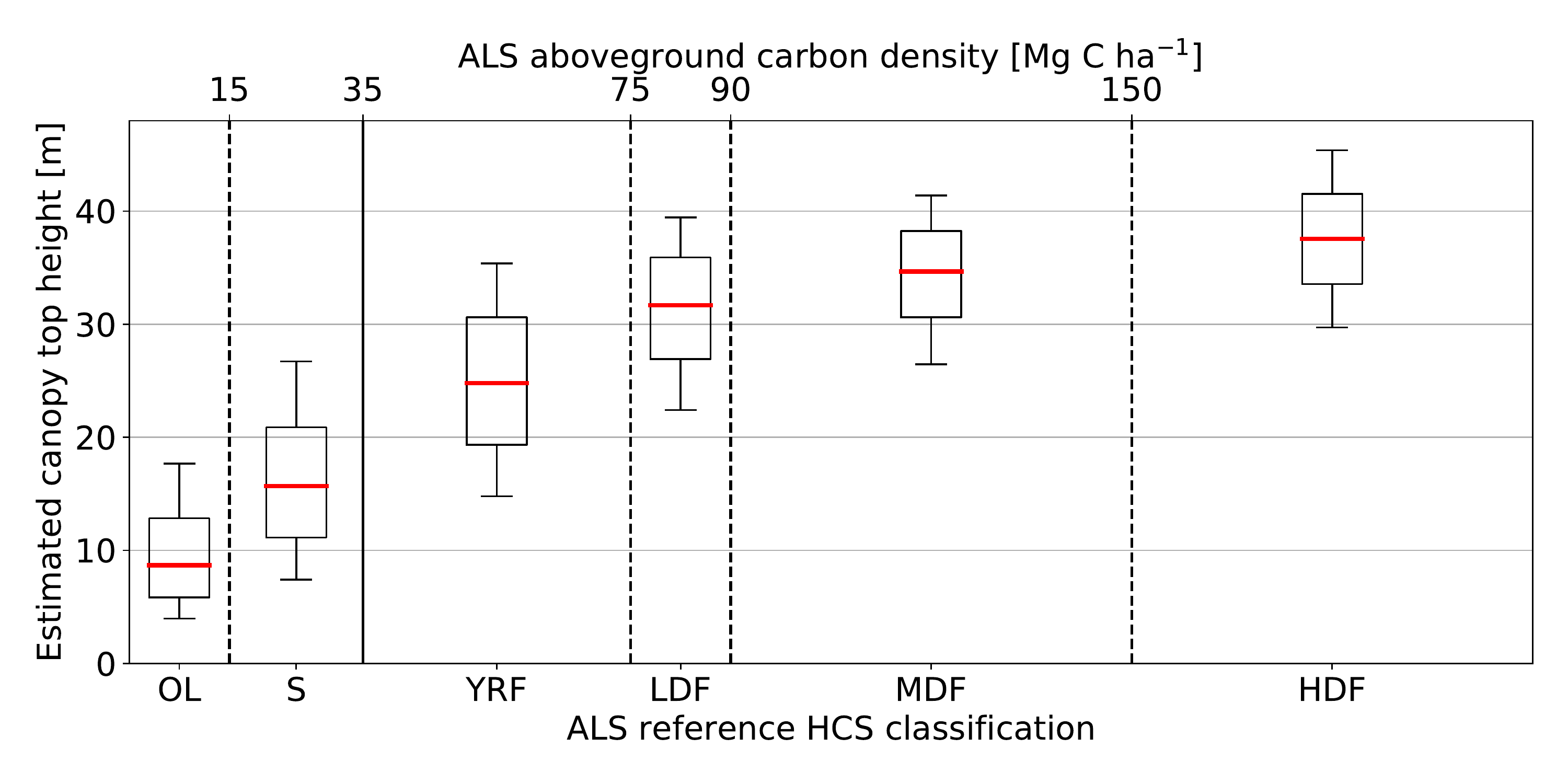} }}%
    \subfloat[]{{ \includegraphics[width=0.4\textwidth]{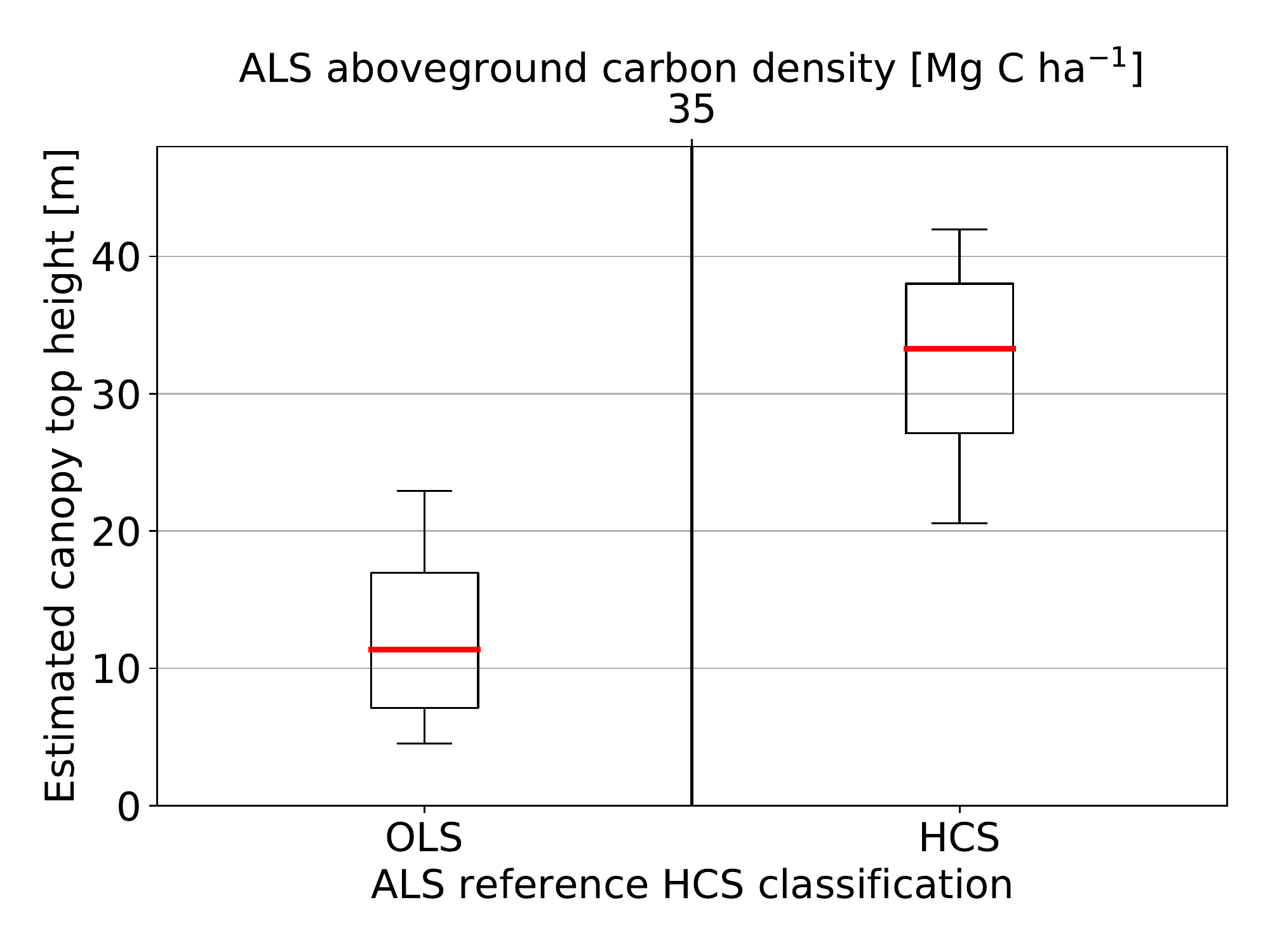} }}
    
    \caption{Distribution of estimated canopy top heights within the reference high carbon stock classification derived from the ALS carbon density. The boxplots depict the median, the quartiles, and the 10th and 90th percentile. a) Six categories: OL: open land, S: scrub, YRF: young regenerating forest, LDF: low density forest, MDF: medium density forest, HDF: high density forest. 
    b) binary classification: OLS: open land \& scrub (OL, S), HCS: high carbon stock (YRF, LDF, MDF, HDF).} 
    \label{fig:canopy_vs_HCS}
\end{figure*}

Qualitatively, we see that the estimated canopy top height depicts the spatial structures of the aboveground carbon density map that is available as a reference from an ALS campaign in Sabah, Borneo Malaysia\cite{asner2018mapped, asner_gregory_p_2021_4549461} (Fig~\ref{fig:als_region}a,~b).
Analyzing the distribution of canopy top height estimates against the HCS classification derived from the ALS carbon density data indicates that canopy top height is predictive of classifying HCS in the Sabah region (Fig.~\ref{fig:canopy_vs_HCS}). While the canopy top height values allow to distinguish well the binary case Fig.~\ref{fig:canopy_vs_HCS}b, the overlap of canopy top height values increases for the finer classification Fig.~\ref{fig:canopy_vs_HCS}a, especially in the high carbon subcategories. 

\begin{figure*}
    \centering
    \subfloat[]{{ \includegraphics[width=0.4\textwidth]{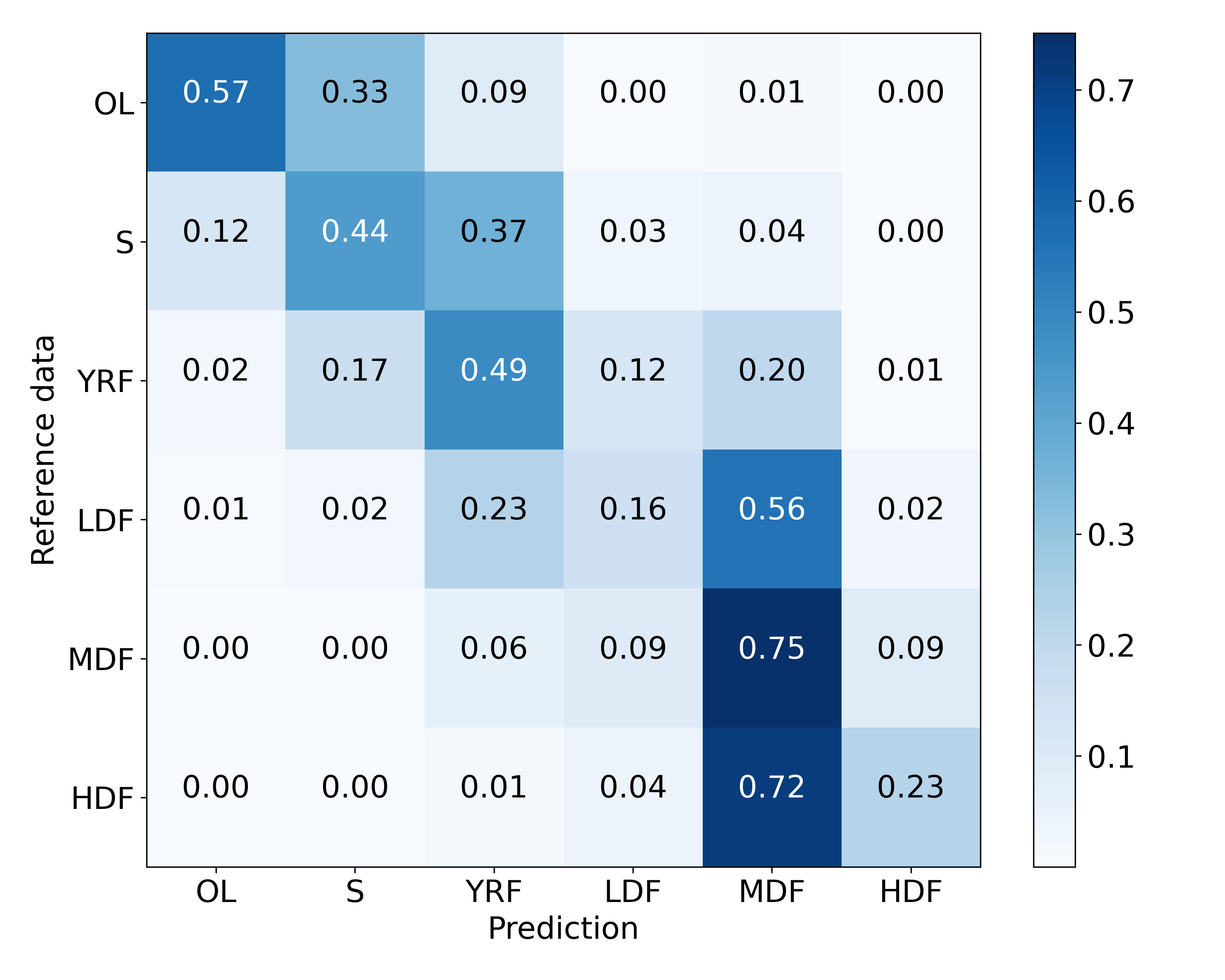} }}%
    \subfloat[]{{ \includegraphics[width=0.4\textwidth]{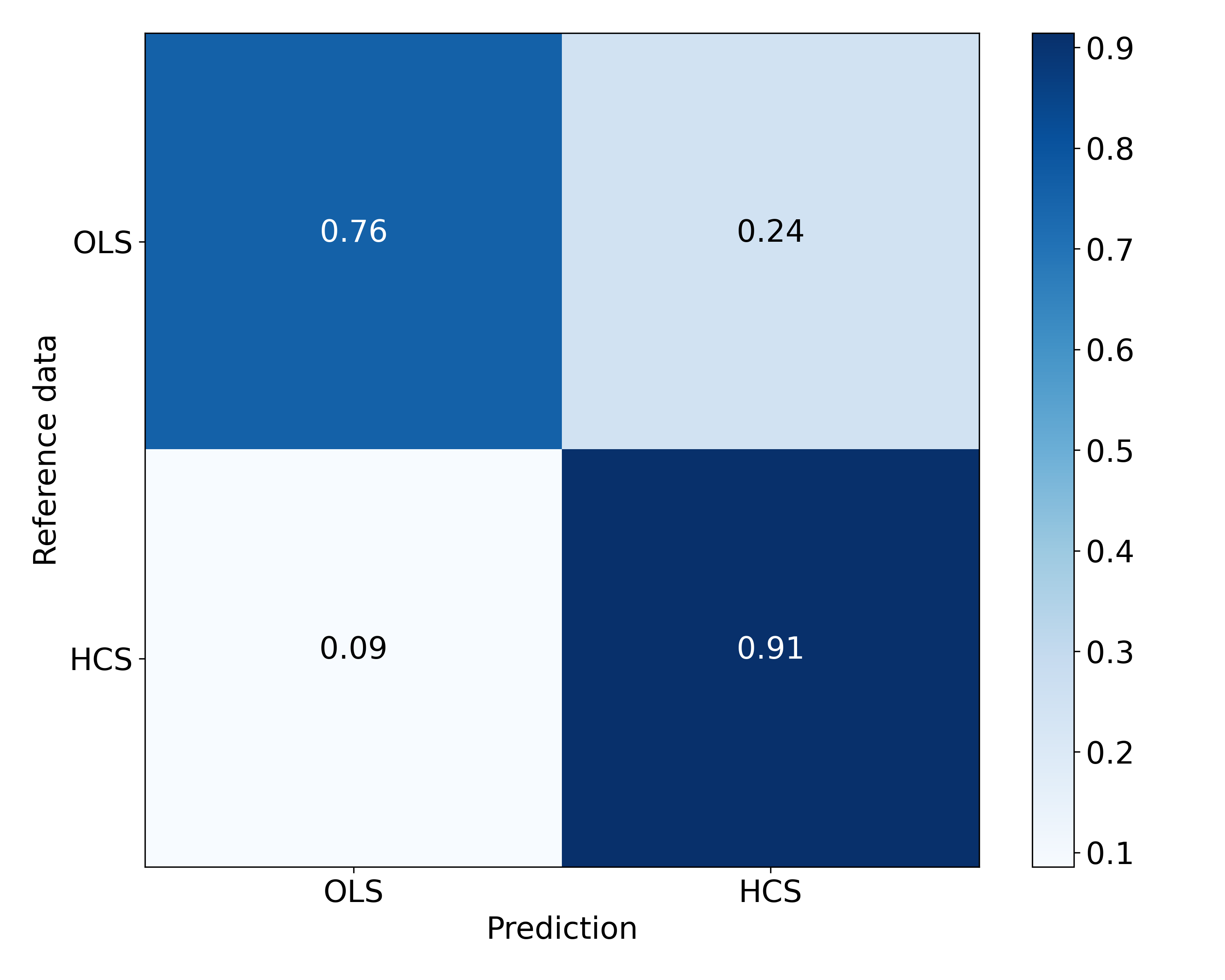} }}
    
    \caption{High carbon stock classification. Confusion matrices for a) Six categories: OL: open land, S: scrub, YRF: young regenerating forest, LDF: low density forest, MDF: medium density forest, HDF: high density forest. 
    b) binary classification: OLS: open land \& scrub (OL, S), HCS: high carbon stock (YRF, LDF, MDF, HDF).} 
    \label{fig:HCS_confusion}
\end{figure*}

The fine-grained definition of the HCSA landscape stratification follows a natural order: open land (OL, <15~Mg~C~ha$^{-1}$), scrub (S, 15-35~Mg~C~ha$^{-1}$), young regenerating forest (YRF, 15-75~Mg~C~ha$^{-1}$), low density forest (LDF, 75-90~Mg~C~ha$^{-1}$), medium density forest (MDF, 90-150~Mg~C~ha$^{-1}$), and high density forest (HDF, >150~Mg~C~ha$^{-1}$)\cite{rosoman2017hcs}. Hence, the critical HCS-threshold is defined at 35~Mg~C~ha$^{-1}$ which separates the latter four high carbon stock (HCS) categories from degraded lands consisting of open land and scrub (abbreviated as OLS). 
Given the natural ordering of the HCS categories and their variable ranges (see also Fig.~\ref{fig:canopy_vs_HCS}a), we prefer to first infer the aboveground carbon density as a regression and then apply the HCSA thresholds, rather than directly map canopy heights to HCS categories.
The ALS calibration site is geographically split and the performance is reported on the test region with an area of 20~km$\times$210~km (Fig.~\ref{fig:als_region}). 
Regressing the carbon density from the dense canopy top height map yields an RMSE of 38.6, a MAE of 27.0, and a ME of 0.9 in Mg~C~ha$^{-1}$. The positive overall bias means that the model overestimates the reference data. The carbon regression saturates around 150~Mg~C~ha$^{-1}$. Consequently, when deriving the HCSA landscape stratification, a significant portion of the high density forest (HDF) is predicted as medium density forest (MDF, see Fig.~\ref{fig:HCS_confusion}a). We observe a slight overestimation of the degraded land subcategories, i.e., 33~\% of open land is classified as scrub and 37~\% is classified as young regenerating forests. In total 19~\% of the young regenerating forests are underestimated to be degraded lands. In the fine-grained classification the overall accuracy is 48~\%, whereby most of the confusion occurs between the adjacent categories (see the block-like structure along the diagonal in the confusion matrix in Fig.~\ref{fig:HCS_confusion}a).  
AT the level of the crucial binary classification (HCS vs.\ OLS) we obtain a higher overall accuracy of 86~\% (Fig.~\ref{fig:HCS_confusion}b). There is a slight bias towards the HCS category: 24~\% of the degraded land is classified as HCS, in contrast only 9~\% of the high carbon stock is misclassified as degraded land.

\subsection*{Large-scale indicative HCS maps}

\begin{figure*}
    \centering
    \subfloat[]{{ \includegraphics[width=0.8\textwidth]{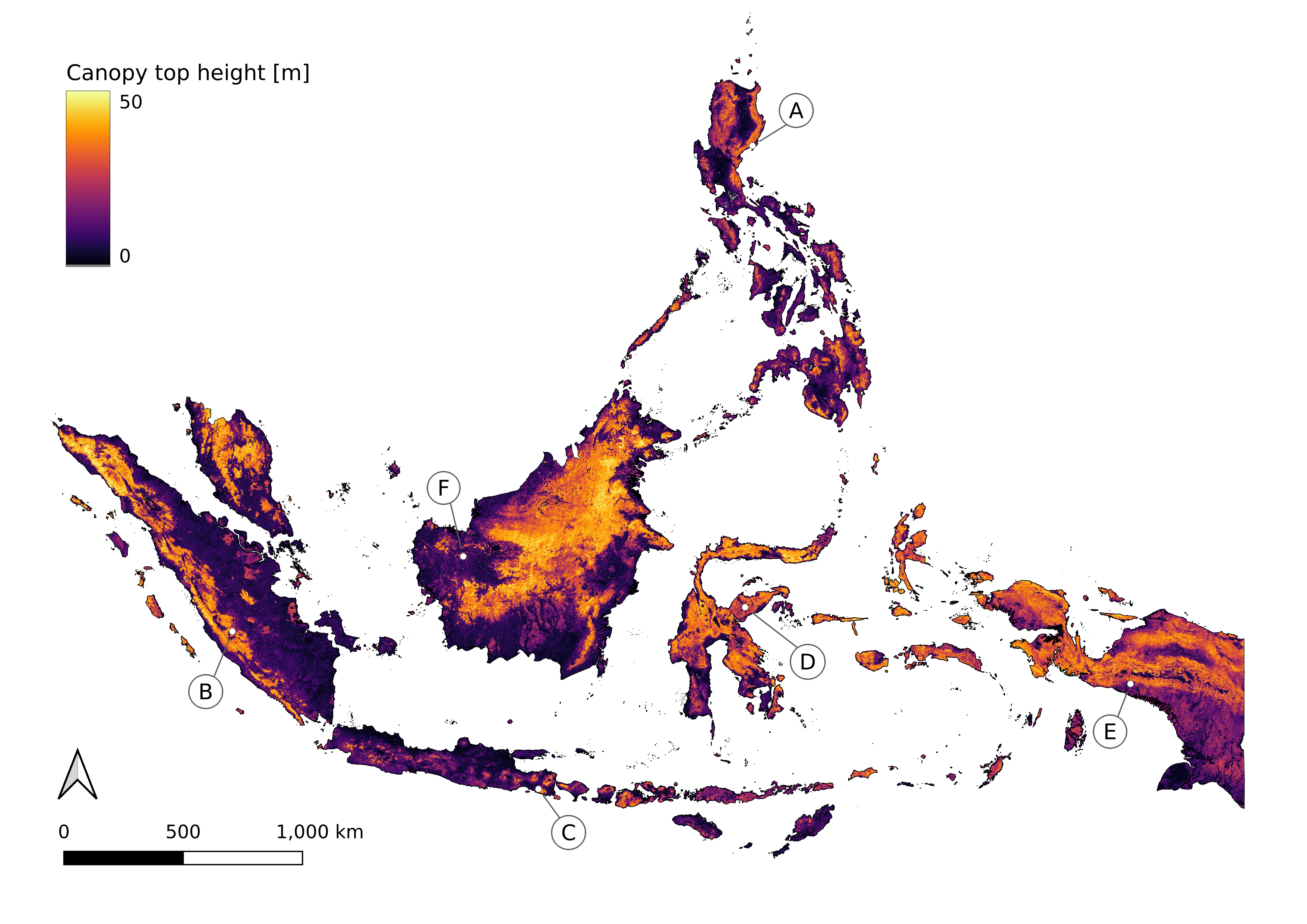} }}
    
    \subfloat[]{{ \includegraphics[width=0.8\textwidth]{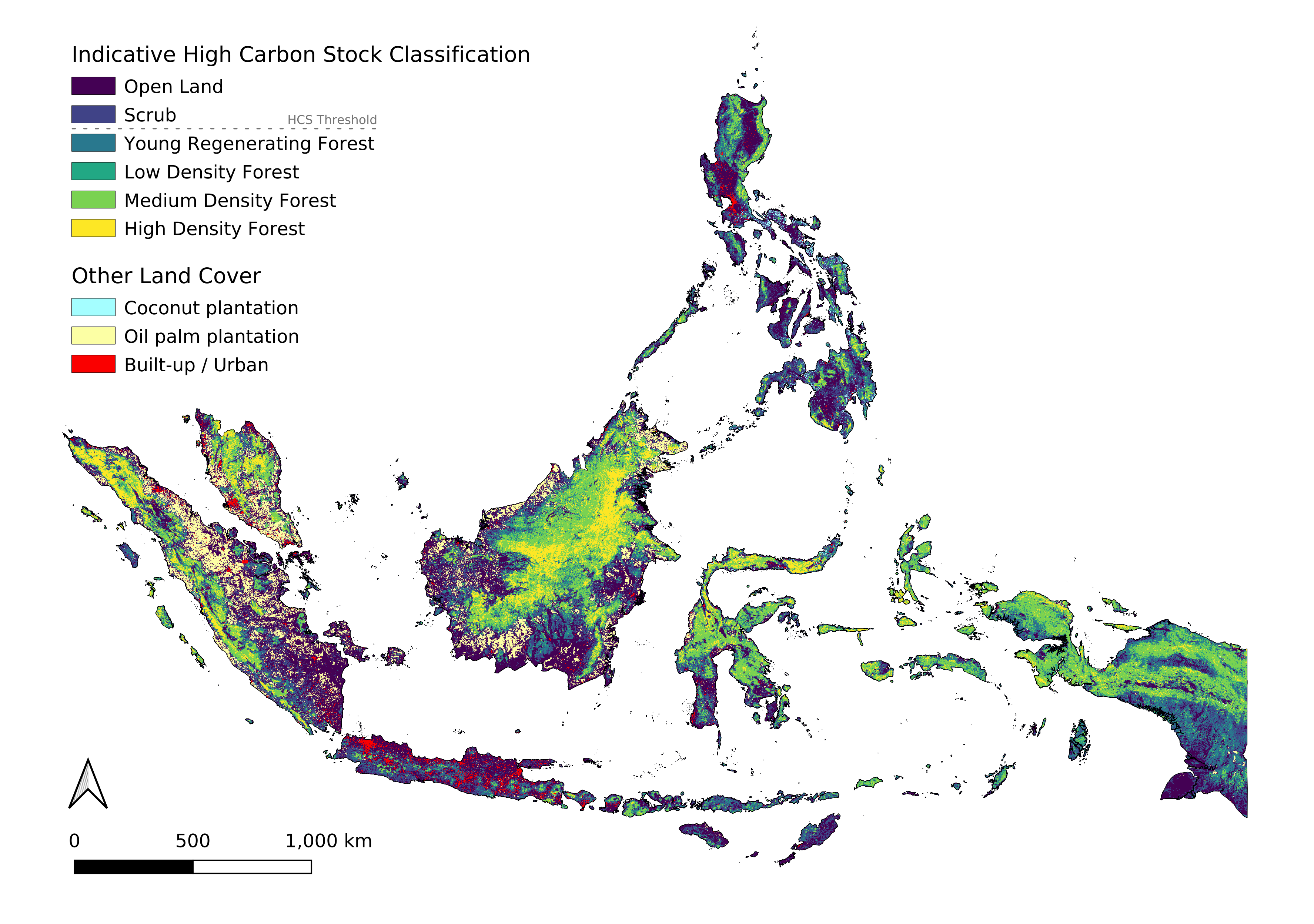} }}
    
    \caption{Maps of Indonesia, Malaysia, and Philippines for the beginning of the year 2021, using images between 1st of September 2020 and 1st of March 2021. a) Canopy top height map estimated from Sentinel-2. The locations A to F are depicted in Fig.~\ref{fig:zoom_qualitative}. b) Indicative high carbon stock map derived from canopy top height. Oil palm for the year 2019 and coconut for 2020 are derived from tree density maps\cite{rodriguez2021mapping,rodriguez2021coconut}. Urban layer from Copernicus Global Land Service 2019\cite{marcel_buchhorn_2020_3939050}.} 
    \label{fig:maps}
\end{figure*}

\begin{figure*}
    \centering
    \includegraphics[width=\textwidth]{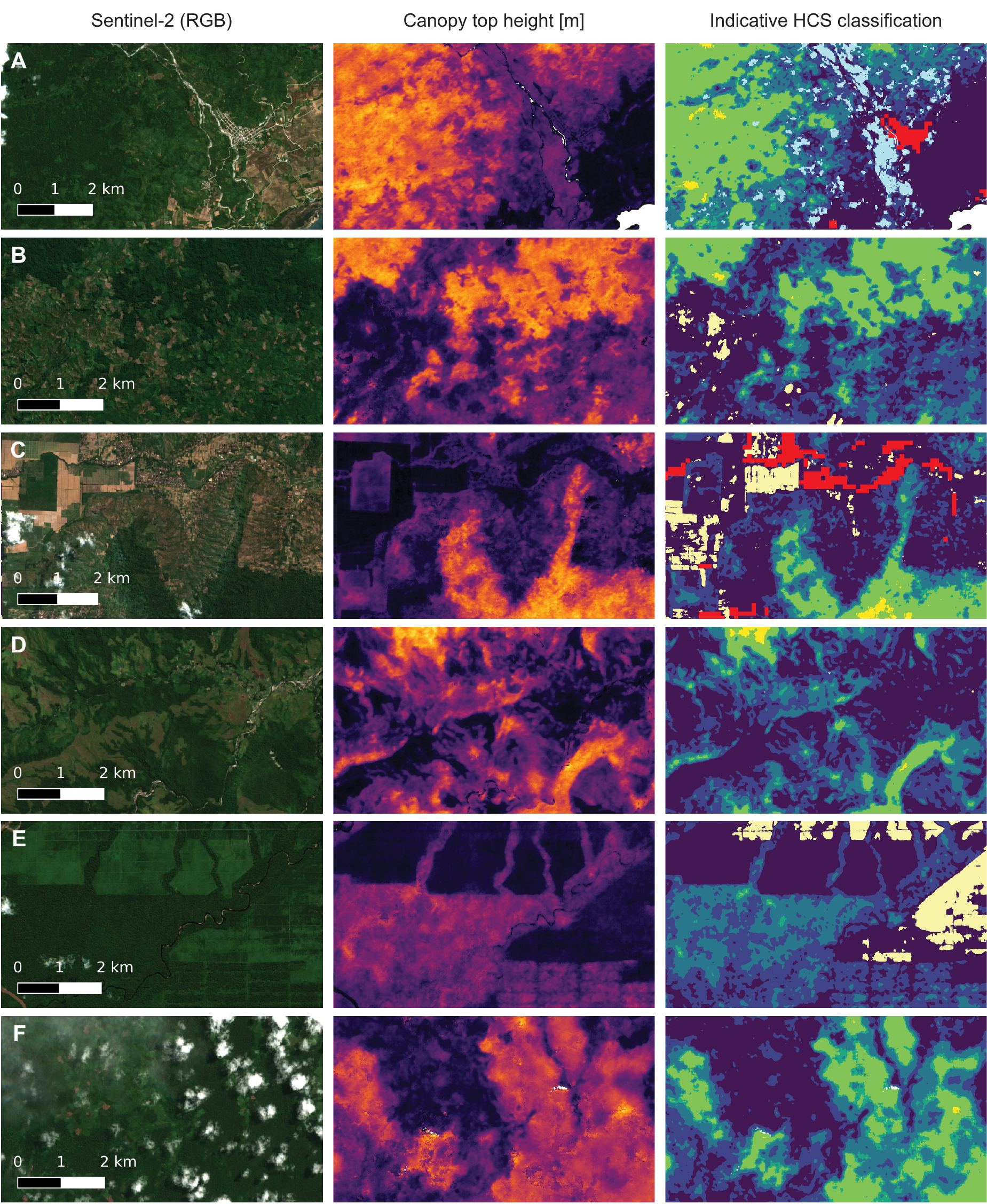}
    \caption{Qualitative results for the locations A to F given in Fig.~\ref{fig:maps}a with the same color coding as in Fig.~\ref{fig:maps}a,b. Left: The Sentinel-2 image with the lowest overall cloud coverage for the respective tile (only RGB true color). Center: Canopy top height estimation. Right: Indicative high carbon stock classification. }
    \label{fig:zoom_qualitative}
\end{figure*}

We computed dense wall-to-wall maps for the beginning of 2021 covering Indonesia, Malaysia, and Philippines (Fig.~\ref{fig:maps}) following the procedure proposed in our previous work\cite{lang2019country}. The region spans a total of 635 Sentinel-2 tiles. For each tile the 10 images with the least cloud coverage between 1st of September 2020 and 1st of March 2021 are processed with the deep CNN. Afterwards, the individual per-image canopy top height predictions are reduced to a single canopy top height map by pixel-wise averaging of all predictions with cloud probability <10~\% (according to the L2A cloud probability mask). This yields a complete map (Fig.~\ref{fig:maps}), even in frequently clouded regions. In rare cases, the cloud masking misses some predictions affected by clouds, which are erroneously included and lead to artifacts in the final canopy height map, for instance Fig.~\ref{fig:zoom_qualitative}F. 
Based on the canopy top height maps the carbon density is regressed with an ensemble of small CNNs and the thresholds defined in the high carbon stock approach are applied to derive the indicative HCS classification (Fig.~\ref{fig:maps}b).
Tree plantations (oil palm, coconut) are explicitly excluded from the HCS classification by overlaying the corresponding masks\cite{rodriguez2021mapping,rodriguez2021coconut}, which were also derived from Sentinel-2. Finally, urban regions are overlaid using the latest existing global land cover product\cite{marcel_buchhorn_2020_3939050}.

To assess the plausibility of the HCS classification in regions where no carbon or HCS reference data is available, six locations (A to F in Fig.~\ref{fig:maps}a) are qualitatively compared with the corresponding Sentinel-2 imagery (Fig.~\ref{fig:zoom_qualitative}). One location in Luzon, Philippines (A), and five locations in Indonesia on the islands: Sumatra (B), Java (C), Sulawesi (D), Papua (E), and Borneo in the province West Kalimantan (F).
Overall, the estimated categories follow the intuitive interpretation of the Sentinel-2 images and spatial features such as forest clearings or smaller forest patches can be recognized. While obvious bare grounds and artificially cleared lands are predicted as low carbon stock, high carbon stock predictions correspond to densely forest areas (dark green with a distinct canopy texture in the Sentinel-2 images).
The model yields plausible low height and carbon predictions even in cases where the per-pixel spectral signature may be ambiguous (e.g., the dark green field in the top left of Fig~\ref{fig:zoom_qualitative}C).
An interesting case is the example from Papua (Fig.~\ref{fig:zoom_qualitative}E) that shows an area undergoing land transformation, namely the extension of oil palm plantations. Distinct clearing patterns can be observed at the top, where the land has already been completely cleared and at the bottom right, where timber harvesting is likely still in process. The resulting maps depict these clearing patterns, including the primary clearing of roads and the thinning of the degraded rectangular forest patches.
The qualitative comparison also confirms that HCS mapping by manual interpretation of satellite images is challenging for human operators.
In particular, human interpretation is limited to, at most, relative ranking. Determining the absolute scale of canopy height or carbon density is practically impossible in a monocular nadir view (e.g. center region in Fig~\ref{fig:zoom_qualitative}D).

We conclude that the relative classification of carbon stock appears plausible in all regions. Having said that, it is not possible to identify certain systematic offsets (say, all forest pixels being assigned one class too low) only with visual inspection. 
To quantitatively evaluate the indicative HCS map, future field campaigns may use the map as guidance to collect targeted validation data in different HCS categories.
Additionally, such targeted field data could also be used as additional reference to re-calibrate the absolute scale of the carbon density, as well as the subsequent HCS classification.

\begin{figure}[tb]
 \centering
 \includegraphics[width=0.9\linewidth]{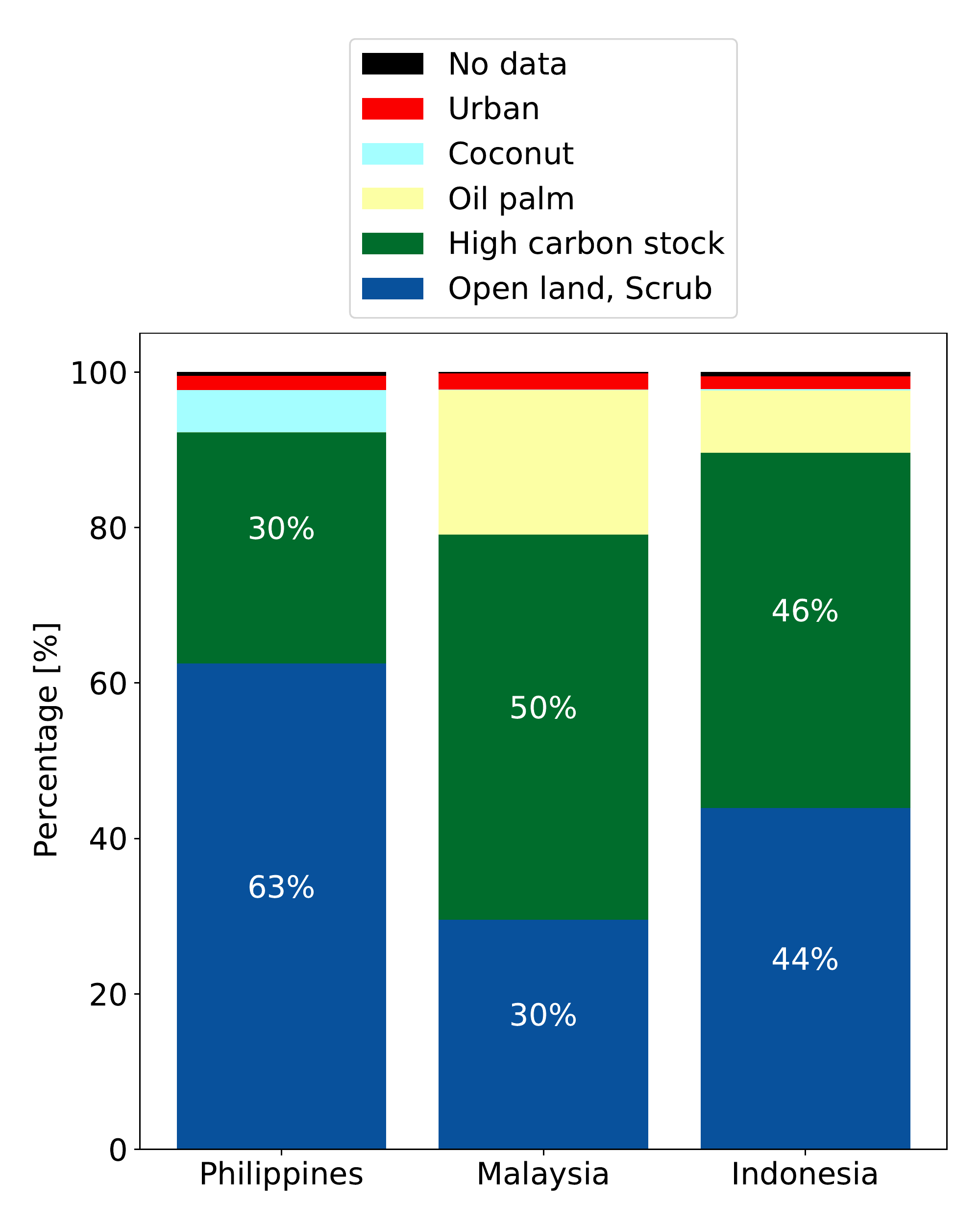}
 \caption{Country-level statistics of mapped land cover categories.}
 \label{fig:stats_country}
\end{figure}

Country-level statistics show that Malaysia has the highest fraction of high carbon stock with 50~\%, followed by Indonesia with 46~\% and Philippines with 30~\% (Fig.~\ref{fig:stats_country}). Malaysia also has, with 19~\%, the higher proportion of plantations (mostly oil palm), compared to Indonesia with 8~\% ,and the Philippines with 5~\% (mostly coconut). In terms of open land and scrub, the Philippines have the highest proportion with 63~\% followed by Indonesia (44~\%) and Malaysia (30~\%).

\begin{figure}
    \centering
    \subfloat[]{{ \includegraphics[width=\linewidth]{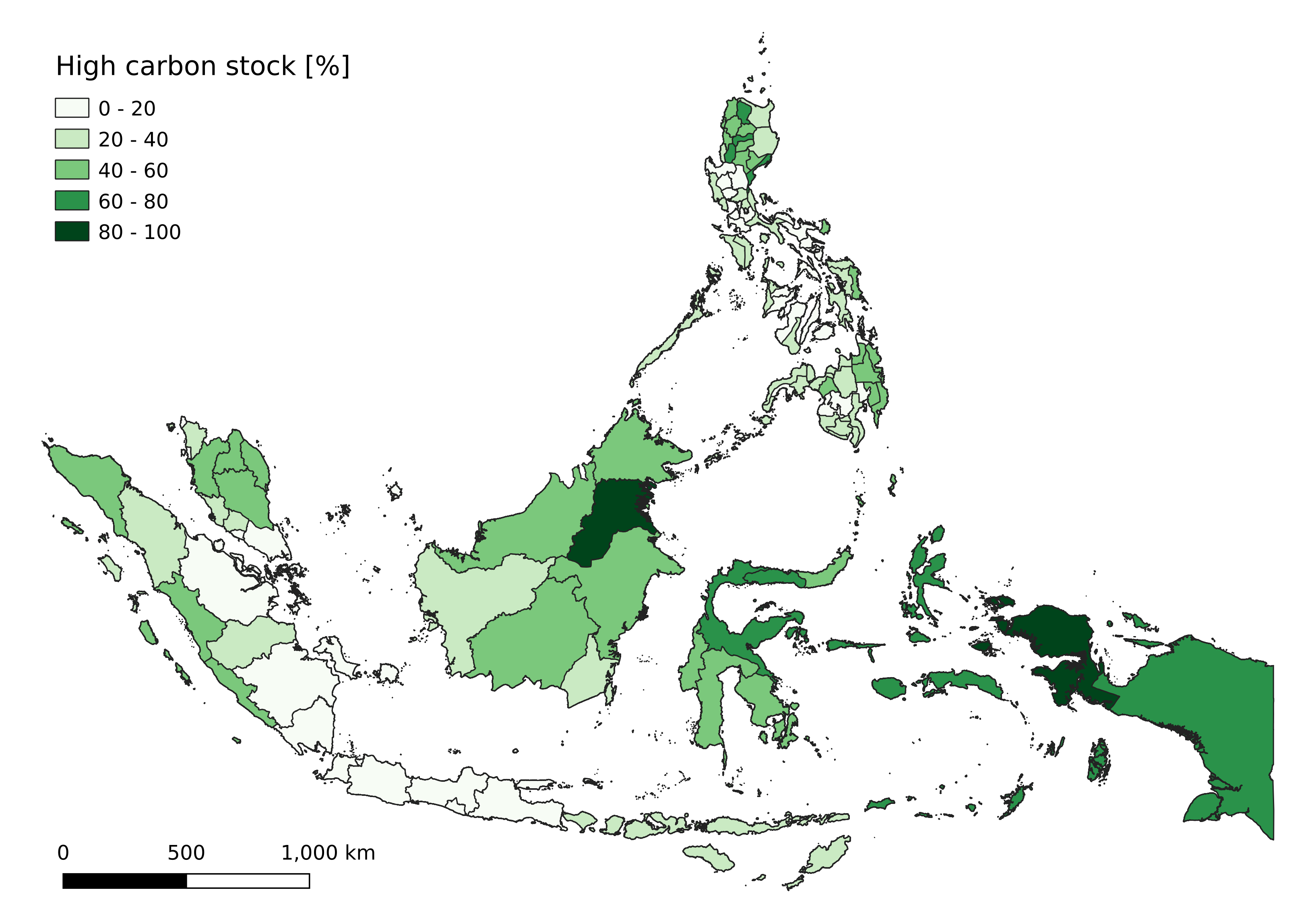} }}
    
    \subfloat[]{{ \includegraphics[width=\linewidth]{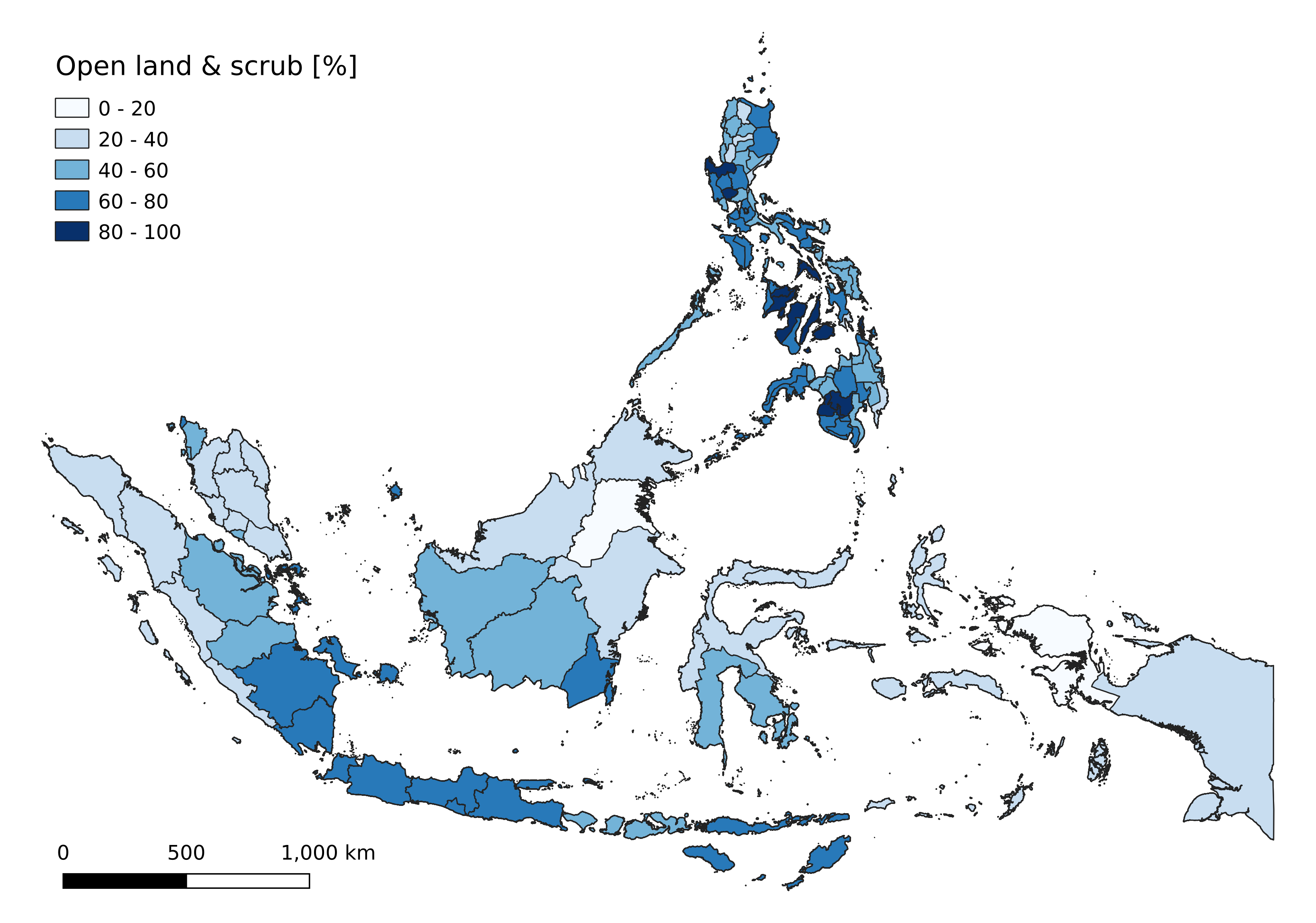} }}
    
    \subfloat[]{{ \includegraphics[width=\linewidth]{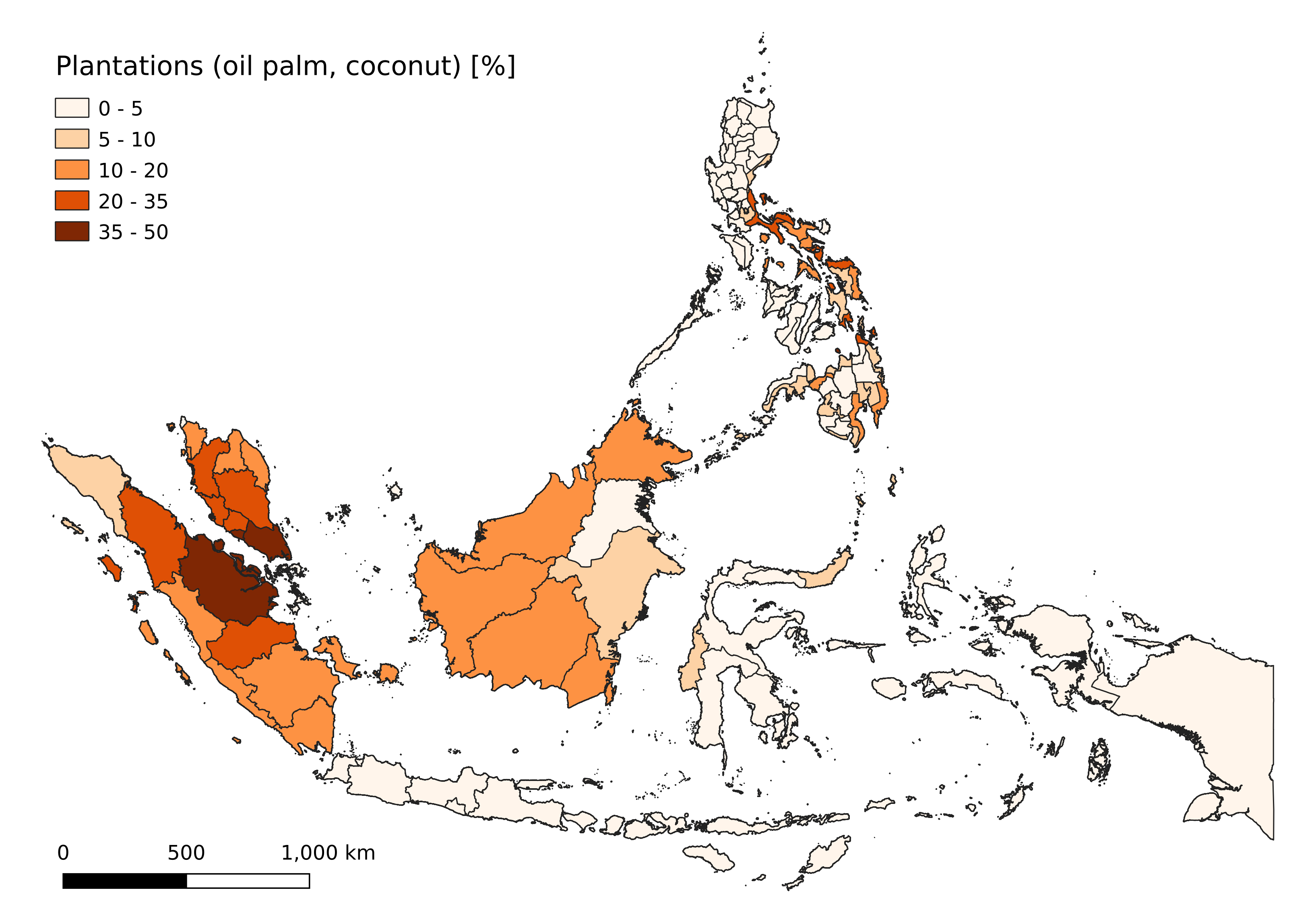} }}
    
    \caption{Distribution of different land covers in each province. a) High carbon stock (HCS), b) Open land \& scrub (OLS), and c) Plantations, i.e., oil palm and coconut.} 
    \label{fig:province_stats}
\end{figure}

At the province level, the two provinces with the largest proportion of high carbon stock per country are Irian Jaya Barat and Kalimantan Utara (Indonesia), Aurora and Apayao (Philippines), and Sarawak and Perak (Malaysia; Fig~\ref{fig:province_stats}a). 
Analogously, the largest proportions of open land and scrub are observed in Guimaras and Siquijor (Philippines), Yogyakarta and Bangka-Belitung (Indonesia), and Perlis and Kedah (Malaysia; Fig~\ref{fig:province_stats}b). Finally, the highest proportion of plantations (oil palm and coconut) were mapped in Johor, Melaka (Malaysia), Riau and Sumatera Utara (Indonesia), and Camiguin and Sorsogon (Philippines; Fig~\ref{fig:province_stats}c).
These preliminary analyses show the potential of the proposed large-scale mapping approach.

\section*{Discussion}
We provide a high-resolution indicative high carbon stock map for Indonesia, Malaysia, and the Philippines.
We show that deep learning offers effective tools to fuse data from recent space missions and create country-wide wall-to-wall maps of canopy top height from optical satellite images. With a learned model of the complex relation between canopy top height and optical image texture (Sentinel-2), we densely interpolate the accurate and well-distributed, but sparse lidar measurements of GEDI. 
Furthermore, we confirm that the resulting canopy top height maps are highly predictive for high carbon stock classification, in Southeast Asia and poentially beyond. 
In our work we have estimated canopy top height and the HCS category for every 10~m$\times$10~m Sentinel-2 pixel, which opens the door towards detailed spatial analyses. We note that the maps, while computed at 10 m GSD, likely have slightly lower effective spatial resolution. Since the neural network was trained with canopy top height data derived from 25~m GEDI footprints, the actual information content of the network output is likely closer to that resolution (as if a GEDI observation with 25~m footprint was placed at the center of every 10~m raster cell).

With an overall accuracy of 86~\% for the binary HCS classification (high carbon stock vs.\ degraded land), the presented approach can already be considered relevant for practical applications. 
We have observed higher confusion within the HCS subcategories, for which there are two possible explanations. First, the canopy height estimates from Sentinel-2 saturate below the maximum tree height of tropical forests, i.e., tall canopies beyond a certain maximum height cannot be distinguished based on texture. Second, in tropical rain forests, biomass growth below the tallest trees is increasingly ambiguous with height\cite{kohler2010towards}. 
Thus, the lack of explicit information about the vertical forest structure, e.g. the canopy density, limits the sensitivity of carbon regression in the high biomass regime.
We note that the saturation does not affect the most important differentiation of the HCS approach, between HCS and degraded lands. Still, improving the fine-grained classification will be important for other potential applications, such as localized carbon accounting.

Why a two step procedure? The main reason is the lack of publicly available carbon reference data. Once a mission like GEDI can provide a large amount of calibrated carbon estimates at the footprint level, the deep convolutional neural network could be trained to estimate wall-to-wall carbon density directly from Sentinel-2 texture, thus ruling out any loss of predictive information along the way.
Until then, our work empirically supports the detour of using deep learning to upscale predictive forest structure variables such as the canopy top height. Deriving such wall-to-wall maps of intermediate predictor variables has another advantage, namely that the HCS classification based on these maps can be calibrated locally. This second step is computationally much cheaper than learning an end-to-end estimation from Sentinel-2 images, and also requires less calibration data, as a model of lower capacity is sufficient to estimate carbon density from canopy top height, compared to the complex perception task to retrieve canopy height from raw optical images. Thus, from the perspective of a local map or decision maker, splitting the procedure into two separate modelling steps has a regularizing effect: they benefit from the global availability of reference data to train the canopy height model, without suffering from the extreme scarcity of up-to-date, global carbon density measurements. 
In terms of decisions based on HCS maps, a two step procedure may help to build trust in the correctness and objectivity of the maps, through the increased transparency of the modelling pipeline. E.g., one can evaluate the consistency of model outputs at different stages, and to some degree quantify the errors introduced in each step. 

In this work, we have used publicly available carbon density data from an airborne lidar campaign in Sabah\cite{asner2018mapped} to calibrate the derivation of carbon stock from the dense canopy height estimates. The generalization of this local calibration to other regions remains to be quantitatively validated with additional reference data. If additional carbon stock data is available, a local re-calibration can likely improve the accuracy.

We refer to the produced HCS map as \emph{indicative}, because the landscape stratification is purely based on the estimated aboveground carbon density. It is important to note that the HCS approach also considers social aspects such as the land use rights of local communities, which cannot be mapped from remote sensing data. Thus, the maps should be seen as a preliminary product that needs to be complemented with administrative and legal clarifications or even field work, but that can also help to efficiently plan these subsequent steps and save resources.

Depending on the application case the indicative HCS map may still miss specific land cover information that must be overlaid to disentangle HCS from other important categories with non-negligible aboveground carbon density.
In our case, only two tree crops were available, whereas other high plantations such as jungle rubber or rubber in mixed forest or agroforestry systems were not explicitly masked and may be included in one of the HCS categories. In contrast, low vegetated areas such as peatlands will, if not explicitly mapped, fall into the OLS category.

As for the relation to ground-based observations, it is important to note that an automated approach is not meant to, and cannot, make field work obsolete in the foreseeable future.
On the contrary, while remote sensing offers excellent tools to densely map large areas that cannot be covered with field observations, it critically relies on field data for calibration and validation. As it lowers the bar to map larger areas and update maps more frequently,
remote sensing arguably even increases the demand for field observations, and the need to better coordinate them regionally and even globally.

In terms of direct ecological impact, our maps and future updates of them may guide the establishment of new reserves and protected areas, both for conservation and to reduce (respectively, offset) carbon emissions. 
Nevertheless, such a "carbon-only" strategy must be complemented by considering further (essential) biodiversity variables to obtain a holistic assessment\cite{pereira2013essential,gardner2012framework}. While biodiversity often correlates with forest density and biomass, also landscapes with low aboveground carbon stock can have high ecological importance\cite{sullivan2017diversity,thomas2013reconciling}. 
Therefore, the HCSA proposes the coupling with the High Conservation Value (HCV) approach\cite{hcsa2020hcsv}.

At a more subtle, but perhaps equally important level the maps may serve as a land use planning tool to develop more sustainably and steer land transformation towards regions where they cause least damage in terms of carbon balance.
The here presented indicative maps could be used in particular by parties that do not have the necessary capacity or resources, such as small and medium scale enterprises (SMEs) and smallholder farmers. Additionally, the maps may be used for commodity source risk mapping by consumer good companies, manufacturers, or commodity traders. 
Moreover, indicative HCS maps are potentially useful to narrow down false alerts in existing deforestation monitoring systems, by focusing on the relevant areas. In that context, future deforestation algorithms could perhaps be designed less conservatively to avoid underestimation of the deforested area\cite{hansen2016humid}.
Another use case could be ecosystem restoration initiatives\cite{chazdon2019restoring,lewis2019regenerate,brancalion2019global,busch2019potential}, which could focus their efforts better on young regenerating forests or degraded lands.

We see this first version of region-wide indicative HCS maps as a first step towards automated, large scale application of the HCS approach. 
We plan to extend the approach presented here to global scale, with priority given to tropical regions covered by the GEDI mission. Future work may involve the quantitative evaluation and re-calibration of carbon stock estimates with the help of additional reference data from different regions. Finally, if the maps are adopted by practitioners, it will be necessary to analyze their impact on practice after a few years.

\section*{Methods}

\subsection*{Data}
This work uses three major data sources: Sentinel-2 optical images, sparse canopy top height estimates from GEDI L1B waveforms\cite{lang2021global}, and a carbon density product from an airborne lidar campaign in Sabah, northern Borneo\cite{asner2018mapped,asner_gregory_p_2021_4549461}. 
In addition, three existing semantic map layers are overlaid as additional filters on the high carbon stock classification: oil palm\cite{rodriguez2021mapping}, coconut\cite{rodriguez2021coconut}, and urban regions\cite{marcel_buchhorn_2020_3939050}.

The Sentinel-2 satellite mission is operated by the European Space Agency (ESA) within the Copernicus program and delivers publicly available multi-spectral images with a revisit time of at least 5 days over the global land masses\footnote{\href{https://sentinel.esa.int/web/sentinel/missions/sentinel-2}{sentinel.esa.int/web/sentinel/missions/sentinel-2} (2021-06-10)}. We use the atmospherically corrected L2A product consisting of 12 bands. All lower resolution bands are up-sampled to match the 10~m resolution of the red, green, blue, and near infrared bands.
The GEDI mission operated by NASA is a space-based lidar system that measures full waveforms (L1B product) that capture profiles of the vertical forest structure within a 25~m footprint on the ground\cite{dubayah2020global}. These measurements are well suited to extract forest structure variables such as the canopy top height, but are sparsely distributed.

Two datasets are constructed for each of the two processing phases, i.e., the dense canopy height regression from the raw Sentinel-2 images and the carbon stock estimation from the dense canopy height map.
The first dataset covers Southeast Asia, where we merge Sentinel-2 images with sparse canopy top height data derived from GEDI L1B waveforms from April to August for the years 2019 and 2020\cite{lang2021global}. Using the Sentinel-2 L2A scene classification product, the canopy top height for non-vegetated areas is set to zero, such that the model is trained to predict zero height for non-vegetated areas.
For each of the 914 Sentinel-2 tiles (approx. 100~km $\times$100~km), we use the image with the lowest overall cloud coverage between May and August 2020 and extract patches of 15$\times$15 pixels centered at each waveform location. Image patches with more than 10~\% cloudy pixels (i.e., pixels with cloud probability >10~\%) are excluded.
For the second phase, the dense canopy top height estimates from Sentinel-2 are paired with a carbon density map from an airborne lidar campaign\cite{asner2018mapped,asner_gregory_p_2021_4549461}. 
This original ALS carbon density product in Mg~C~ha$^{-1}$ has a resolution of 30~m and is bilinearly re-sampled to match the 10~m (Sentinel-2) raster of the dense canopy height map.
Since the ALS data has been recorded in 2016, we produce a canopy top height map for this region using Sentinel-2 images from the year 2017 to reduce temporal discrepancies. Plantations are masked out using the 2017 oil palm density map\cite{rodriguez2021mapping}.

To mask urban (built-up) regions, we fall back to the latest existing map layer for the year 2019 provided by the Copernicus Global Land Service\cite{marcel_buchhorn_2020_3939050}, which has a resolution of 100~m. 
In addition, two types of tall crop plantations are explicitly masked, oil palm and coconut. Semantic binary maps of oil palm plantations are created by thresholding an existing oil palm tree density map for the year 2019\cite{rodriguez2021mapping}. The same methodology has been applied for coconut plantations for the year 2020\cite{rodriguez2021coconut}. The density thresholds, in trees per Sentinel-2 pixel, are empirically set to >0.2 for oil palm and to >0.4 for coconut.

\subsection*{Combining GEDI and Sentinel-2 for dense canopy top height mapping}
A deep fully convolutional neural network is trained, adapting our previous work where canopy height retrieval from Sentinel-2 was learned regionally from airborne lidar reference data\cite{lang2019country}. Here, we use GEDI derived canopy top height estimates as the reference data to train the same architecture. The difference is that the GEDI reference data is sparse and distributed. Since the network is fully convolutional, an output is computed for every input pixel, but at training time only the pixels with a valid reference height are used to compute the loss and update the network parameters. We use the mean squared error (MSE) as the loss function, but also report the mean absolute error (MAE) and the mean error (ME), where a negative ME (bias) means that the prediction is lower than the reference.
To train and validate a single model for Southeast Asia, we split the dataset at the tile level into two subsets, holding out a random set of 92 tiles (10~\%) for validation, with a total of $2.7\times10^6$ validation samples. The remaining 822 tiles are used for training, which corresponds to $24.6\times10^6$ training samples. The CNN is trained for 725,000 iterations with 64 image patches per iteration using the ADAM optimization scheme\cite{kingma2015adam}, starting with a base learning rate of 0.0001. 

\subsection*{Estimating carbon stock from canopy top height}
The computed dense canopy top height is translated into carbon density by learning an ensemble of five shallower fully convolutional networks. The CNNs take the dense canopy height map derived from Sentinel-2 as an input and predict carbon density for every 10~m$\times$10~m input raster cell. The predictions of the five individual models are averaged and used as the ensemble prediction\cite{lakshminarayanan2016simple}. 
The CNNs consist of a stack of convolutional layers with 3x3 filter kernels. With a receptive field of 15$\times$15 pixels the model can also extract canopy height patterns from the pixel neighbourhood and is not restricted to treating each pixel in isolation. Each model is trained individually for 100 epochs using ADAM with learning rate 0.0001 on the 170~km$\times$210~km training region (Fig~\ref{fig:als_region}), by minimising the negative Gaussian log-likelihood. Since monitoring validation region did not indicate any overfitting, we refrain from early stopping and use the final model after the last epoch for evaluation on the test region. 
The resulting carbon density estimates are thresholded using the suggested values defined by the HCSA Toolkit Module 4\cite{rosoman2017hcs} to obtain the high carbon stock classification.
Since previous works suggest that aboveground carbon density and canopy height are related by a power-law\cite{kohler2010towards,asner2014mapping}, we include a power-law transformation with trainable parameters as the first layer of the CNN. However, we did not observe any notable difference in performance, it appears that such a relationship can be approximated with the standard CNN layers, if applicable.

\bibliography{bibliography}

\section*{Acknowledgements}
The project received funding  from  Barry  Callebaut  Sourcing  AG, as part of a Research Project Agreement. We thank Grant Rosoman for his inputs and comments.




\end{document}